\documentclass[10pt,twocolumn,letterpaper]{article}

\usepackage{cvpr}
\usepackage{times}
\usepackage{epsfig}
\usepackage{graphicx}
\usepackage{amsmath}
\usepackage{amssymb}
\usepackage{subcaption}
\usepackage{comment}
\usepackage{ulem}

\usepackage[symbol]{footmisc}

\usepackage{soul,xcolor}
\setstcolor{green}

\usepackage[pagebackref=true,breaklinks=true,letterpaper=true,colorlinks,bookmarks=false]{hyperref}

\usepackage[capitalise]{cleveref}

\cvprfinalcopy 

\def\cvprPaperID{****} 

\ifcvprfinal\fi
\begin{document}
\title{Enhancing Facial Data Diversity with Style-based Face Aging}

\author{Markos Georgopoulos $^{1}$\thanks{co-first authorship.}, \quad James Oldfield$^{2}$\footnotemark[1], \quad Mihalis A. Nicolaou$^{2}$, \\ \quad Yannis Panagakis$^{3}$, \quad Maja Pantic$^{1}$ \\ 
{\textsuperscript{1} Department of Computing, Imperial College London, United Kingdom}\\
{\textsuperscript{2} Computation-based Science and Technology Research Center, The Cyprus Institute}\\
{\textsuperscript{3} Department of Informatics and Telecommunications, University of Athens, Greece}\\
{\texttt{\{m.georgopoulos, i.panagakis, m.pantic\}@imperial.ac.uk}}\\
{\texttt{\{j.oldfield, m.nicolaou\}@cyi.ac.cy}}}

\maketitle
\thispagestyle{empty}

\begin{abstract}
   \looseness-1A significant limiting factor in training fair classifiers relates to the presence of dataset bias.  In particular, face datasets are typically biased in terms of attributes such as gender, age, and race.  If not mitigated, bias leads to algorithms that exhibit unfair behaviour towards such groups.  In this work, we address the problem of increasing the diversity of face datasets with respect to age.  Concretely, we propose a novel, generative style-based architecture for data augmentation that captures fine-grained aging patterns by conditioning on multi-resolution age-discriminative representations.  By evaluating on several age-annotated datasets in both single- and cross-database experiments, we show that the proposed method outperforms state-of-the-art algorithms for age transfer, {especially in the case of age groups that lie in the tails of the label distribution}.  We further show significantly increased diversity in the augmented datasets, outperforming all compared methods according to established metrics.
\end{abstract}

\section{Introduction}

Face analysis technology has penetrated in our daily lives by becoming a core component in human-machine interaction while it plays an increasingly important role in the decision-making of several processes involving humans.
Hence, it is crucial for facial analytics systems to produce objective and fair results. However, despite the transformative capabilities of deep learning models in face analysis, there are persistent issues. For instance, bias in facial data occurs when 
there is unequal representation of protected attributes such as age, gender, skin color or ethnicity. Computer vision algorithms trained on datasets encoding such biases can result in biased performance across 
vulnerable or underrepresented groups \cite{torralba2011unbiased}.

In this paper, we focus on \textit{age bias}, which results from the scarcity of available images that depict very old or young faces, and is considered  one of the most common biases in face analysis. 
Indicatively, most widely used age-annotated datasets (e.g., MORPH \cite{morph}, CACD \cite{cacd}, IMDB \cite{dexRothe-ICCVW-2015}, FG-NET \cite{FGNET}, AgeDB \cite{agedb}, AFAD \cite{afad} have significantly imbalanced age distributions; for instance, 87\% of the samples in FG-NET are younger than 30 years old.

Traditional techniques to handle class-imbalance, such as image transformations and augmentation are of limited benefit as they are not able to produce realistic approximations to the underlying data distribution across ages. A more recent alternative  is the use of generative models to augment existing datasets and augment the underrepresented age classes.
However, such approaches 
do not address the inherent problem of dealing with small and unbalanced training sets. Indeed, early works on age progression (e.g., \cite{ramanathan2006modeling, suo2009compositional, kemelmacher2014illumination}) were not able to produce photorealistic results due to the simplicity of the models and the lack of available training data. Similarly, more recent GAN-based frameworks (e.g., \cite{ipcgan,caae}) fail to synthesize faces of extreme age (i.e., very young or very old). Moreover, conditioning facial synthesis on a single label ignores the intra-class diversity of each age class and collapses them to a single aging pattern. As a result, recent GAN-based methods produce a single age progression/regression per face, which in most cases is biased according to the age distribution of the training set. These shortcomings deem such methods ineffective for diversity-enhancing data augmentation, since training a model on a biased synthetic dataset would still result in algorithmic bias.

\begin{figure*}
    \centering
    \includegraphics[width=1.0\textwidth]{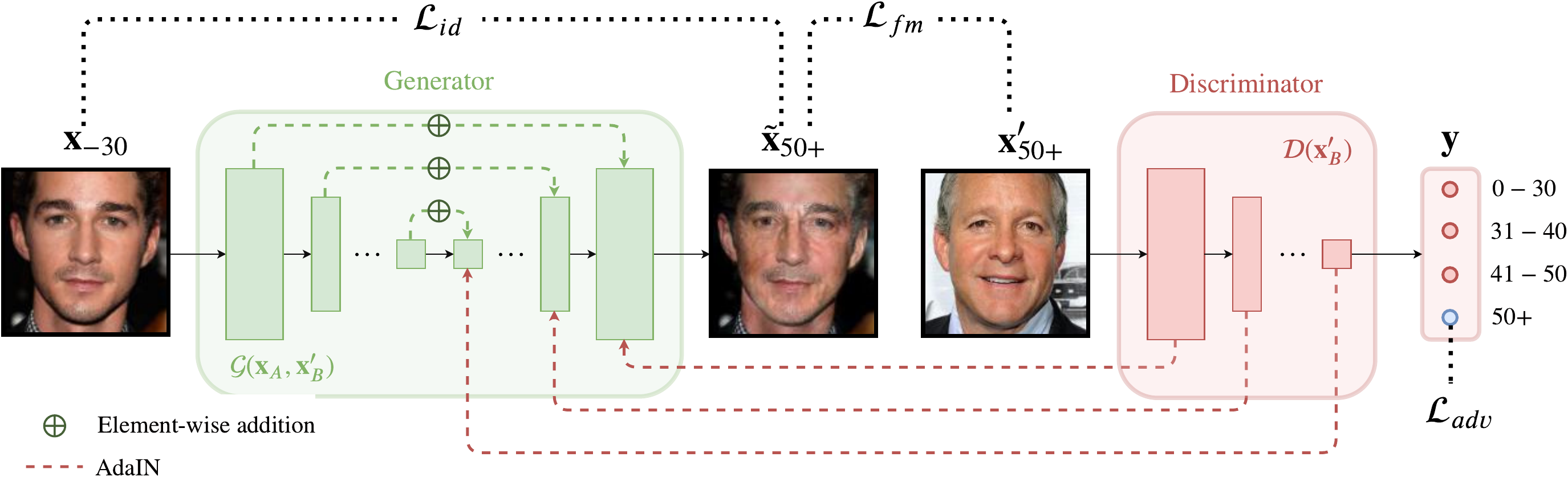}
    
        \caption{\looseness-0Overview of our method for face aging. First, a real image $\mathbf{x'}_{50+}$ of the target age group $50+$ is fed into the discriminator (right). The channel-wise mean and standard deviation of the resulting activations at each layer are used to modulate the statistics of the corresponding feature maps in 
        the generator. Shown here is the aging of image $\mathbf{x}_{-30}$ (left) of age group $0-30$ into $\mathbf{\tilde{x}}_{50+}$ of age group $50+$. In minimizing $\mathcal{L}_{adv}$ the discriminator outputs high probability of the real image being of the expected age group--thus naturally building age-discriminative representations. A feature-matching loss $\mathcal{L}_{fm}$ is used to encourage the generator to match the precise aging patterns of the target image. The identity-preservation loss $\mathcal{L}_{id}$  is applied to ensure that the identity-specific features of the input image are preserved.}
   
    \label{fig:overview}
\end{figure*}

In this work, rather than generating images by conditioning on labels, we propose a style-based age transfer framework that is tailored to the task of diversity-enhancing facial data augmentation.
In particular, we introduce a style-transfer approach that allows the model to synthesize diverse aging patterns based on the choice of target image. Furthermore, by conditioning the generation on extreme targets, e.g., very old-looking faces, we are able to generate sufficiently old/young-looking faces despite the lack of support in the training set. 
Concretely:
\begin{itemize}
    \item We introduce a novel GAN that is able to synthesize an aged/rejuvenated face by conditioning on the age-related style of a target face at multiple scales. 
    
    \item We showcase the aging accuracy of our model through qualitative and quantitative experiments. The proposed model is compared against strong baselines in \cref{section:qualitative-results,section:aging-accuracy}.
    
    \item The ability of our model to diversify the age distribution of a biased dataset is investigated in \cref{section:diversity}. We propose to quantify this by using the metrics introduced in \cite{dif-merler2019diversity}.
    
    \item Lastly, we showcase the ability of our model to synthesize diverse aging patterns in \cref{section:intraclass}. We show that by conditioning the generation on a target image rather than a target class, the proposed framework is able to generate faces of large intra-class diversity.
\end{itemize}{}


\section{Related work}
\paragraph{Generative adversarial networks} Generative adversarial networks constitute the state-of-the-art approach for generative modeling and have been successfully applied to tasks such as image generation \cite{dcganradford2015unsupervised,stylegan}, image captioning \cite{naturalcaptionDai_2017,captionChen_2019}, super-resolution \cite{superresolutionLedig_2017,wganresolution-chen2017face}, and image translation \cite{starganChoi_2018,munit,ugat,pix2pixIsola_2017,cycleganZhu_2017}. 
In the setting of image translation, that is most closely related to this work, the goal is to learn a mapping from an image domain $X$ to a domain $Y$.  
Successful approaches have been proposed, both for
the paired \cite{pix2pixIsola_2017} and unpaired \cite{cycleganZhu_2017,starganChoi_2018} image-to-image translation settings. Recent state-of-the-art draws inspiration directly from the style transfer literature, modulating style content explicitly at different layers with the Adaptive Instance Normalization \cite{adain} (AdaIN) operation \cite{munit, ugat}. Our framework draws inspiration from the GAN-based style transfer literature.

\paragraph{Style Transfer} The seminal work of Gatys et al. \cite{gatys} showed that the global style of an image is captured in the gram matrix of vectorised feature maps in a CNN and can be transferred independently of the image content. Subsequent works focused on improving the speed \cite{Johnson_2016,Li_2016,ulyanov2016texture} or flexibility \cite{Gu2018ArbitraryST, universal-wct,Sanakoyeu_2018,adain,batchnorm-domain} of style transfer.
More recently, research into the interpretation of its success has been ongoing. In particular, Li et al. showed in \cite{demyst} that the matching of gram matrices can be viewed as a form of distribution alignment, while others treat style transfer formally as the optimal transport problem \cite{closed-form-optimal-transport,mroueh2019wasserstein}. 
In this work, we focus on the age-discriminative style of the image and use style-transfer to perform face aging.

\paragraph{Face aging} Face aging has been studied extensively in the fields of anthropometry and computer graphics, before transitioning to computer vision. The reader is referred to \cite{fu2010age, ramanathan2009age, georgopoulos2018modeling} for comprehensive surveys on facial age progression. Early works focused on modeling biologically inspired mechanical transformations and facial anatomy \cite{todd1980perception, ramanathan2006modeling, suo2009compositional}. These physical model-based methods were computationally expensive and could not generalize well due to the constraints of the models. Later data-driven approaches would learn a mapping between age class prototypes (e.g., class mean) \cite{burt1995perception, kemelmacher2014illumination}. These age progression  methods suffered from the loss of identity information resulting in unrealistic aging results. With the establishment of deep learning techniques recurrent \cite{wang2016recurrent} and GAN-based \cite{ipcgan, caae, psdganYang_2018} architectures have been  utilized to perform face aging. In \cite{caae}, a conditional adversarial autoencoder (CAAE) is proposed and face aging is performed by traversing a low-dimensional manifold. Wang et al. \cite{ipcgan} utilize pre-trained networks to preserve identity and achieve aging accuracy. Similarly, Yang et al. \cite{psdganYang_2018} use age features from a pre-trained network in the discriminator. In contrast to such existing methods, our framework is not only trained end-to-end without auxiliary classifiers, but is also able to transfer diverse aging features. By leveraging target faces of extreme age, the proposed framework is able to synthesize the aging patterns of both the very old and young faces.
This allows for the use of our method as a data augmentation tool for mitigating bias in datasets.

\paragraph{Bias mitigation} Different approaches have been proposed to mitigate bias from a model. Inspired by domain adaptation, Alvi et al. \cite{alvi2018turning} proposed a joint learning and un-learning framework, while Kim et al. \cite{kim2019learning} minimize the mutual information between the network embedding and bias information. The use of generative models for fair data augmentation has been investigated in \cite{sattigeri2018fairness, fairganXu_2018, quadrianto2018discovering}.
In order to generate a complete dataset that can be used to train a fair classifier, GAN-based methods with fairness constraints were proposed in \cite{sattigeri2018fairness, fairganXu_2018}. On the other hand, Quadrianto et al. \cite{quadrianto2018discovering} introduced an autoencoder that removes sensitive attribute information from the data. Contrary to this work, these generative methods focus mainly on generating data that can be used to train a fair classifier, but are not necessarily naturalistic (e.g., gender-less faces in \cite{quadrianto2018discovering}).

\section{Methodology}
\looseness-1In this section, we describe the proposed methodology that is focused on enhancing the diversity of a given face dataset with respect to age.  Inspired by recent progress in style transfer, the proposed  architecture is specifically designed to provide fine-grained control over  aging patterns.  This is achieved 
by conditioning the autoencoder-based  generator  on multi-resolution age-discriminative representations.  In this way, we further   relax the rigid assumption of dependence on a single class label for an age group, unlike previous works \cite{ipcgan,caae, psdganYang_2018}. We posit that by employing the proposed approach, we can both capture fine-grained aging patterns as well as accurately synthesize realistic samples that lie in the tails of the dataset distribution, thus significantly increasing dataset diversity.   The remainder of this section is structured as follows. In \cref{sec:meth:propframe} we introduce our modeling choices for the generator and discriminator networks, while in \cref{sec:meth:trainobj} we describe the proposed training objective.  An overview of the proposed method is visualized in \cref{fig:overview}.

\subsection{Proposed framework}
\label{sec:meth:propframe}

\textbf{Style-conditioned Generator}:   
We adopt an autoencoder-based architecture for the generator $\mathcal{G}$, that is  trained to translate an input face image $\mathbf{x}_A$ of age $A$ to a synthesized image age $\mathbf{\tilde{x}}_B$ of age $B$, using the aging patterns of a target image $\mathbf{x}'_{B}$. The target age style is extracted from $\mathbf{x}'_{B}$ by the disciminator network $\mathcal{D}$. In particular, the age information at different scales is obtained from the first and second order moments of the features at different layers of $\mathcal{D}(\mathbf{x}'_{B})$. The age-discriminative style is then injected into the decoder of the generator using AdaIN. Following the paradigm of \cite{unetRonneberger_2015}, we utilize skip connections between the layers of the encoder and the decoder, to mitigate training instability issues. 

\textbf{Discriminator}: The discriminator of the proposed framework is trained to distinguish between real and fake images of each class. To this end, we adopt the multi-task discriminator of \cite{fewshot}. The resulting network captures features that represent both the ``realness'' of the faces as well as their age. The architecture of $\mathcal{D}$ is a mirrored decoder of the generator, in order to maintain correspondence between the features at different scales (i.e., layers).

\subsection{Training objective}
\label{sec:meth:trainobj}
The following objective function of the model is comprised of three parts, namely: the adversarial loss, the reconstruction loss, and the identity preservation loss.

\textbf{Adversarial loss}: For our framework to be able to synthesize photorealistic images, we train $\mathcal{G}$ and $\mathcal{D}$ using an adversarial loss. Given an input image $\mathbf{x}_A$, a target image $\mathbf{x}'_B$, and the age-progressed/regressed output of the generator $\mathbf{\tilde{x}}_B = \mathcal{G}(\mathbf{x}_A, \mathbf{x}'_B)$, the adversarial loss is calculated as follows:
\begin{align}
    \mathcal{L}_{adv} =
        &\mathbb{E}_{\mathbf{x}_{A}}\big[ \log\mathcal{D}(\mathbf{x}_A) \big] + \nonumber \\
        &\mathbb{E}_{\mathbf{x}_A, \mathbf{x'}_B}\big[ \log(1-\mathcal{D}\big(\mathcal{G}(\mathbf{x}_A, \mathbf{x'}_B)\big) \big],
\end{align}

In the typical GAN setting, the generator tries to minimize $\mathcal{L}_{adv}$, while the discriminator tries to maximize it. However, in order to maintain the diversity in aging patterns among different target faces, we train the generator using a feature-matching loss \cite{improved-gans}:
\begin{equation}
  \mathcal{L}_{fm} = \mathbb{E}_{\mathbf{x}_A,\mathbf{x'}_B} \big[ \parallel \mathcal{D}(\mathbf{x'}_B) - \mathcal{D}\big(\mathcal{G}(\mathbf{x}_A, \mathbf{x'}_B)\big) \parallel_2^2 \big].
\end{equation}

\textbf{Reconstruction loss}: In order to ensure that $\mathcal{G}$ preserves the content of the input image, we minimize a reconstruction loss. That is, we enforce cycle consistency by transforming an input image $\mathbf{x}_A$ using a target $\mathbf{x}'_B$ and subsequently transforming back to the original by using $\mathbf{x}_A$ as the target. Concretely:
\begin{equation}
  \mathcal{L}_{rec} = \mathbb{E}_{\mathbf{x}_A,\mathbf{x'}_B} \big[ \parallel \mathbf{x}_A - \mathcal{G}\big(  \mathcal{G}(\mathbf{x}_A, \mathbf{x'}_B), \mathbf{x}_A\big) \parallel_1 \big].
\end{equation}

\textbf{Identity preservation}: Besides maintaining the original content, it is vital for the task of age progression to maintain the person-specific high-frequency details of the input. Therefore, we minimize a pixel-wise $L1$ loss between the input the output of $\mathcal{G}$:

\begin{equation}
  \mathcal{L}_{id} = \mathbb{E}_{\mathbf{x}_A,\mathbf{x'}_B} \big[ \parallel \mathbf{x}_A - \mathcal{G}(\mathbf{x}_A, \mathbf{x'}_B) \parallel_1 \big].
\end{equation}

\textbf{Full objective}: Based on the above, $\mathcal{G}$ and $\mathcal{D}$ are trained to minimize the following composite loss functions:
\begin{align}
    \mathcal{L}_{D} &=-\mathcal{L}_{adv} \\ 
    \mathcal{L}_{G} &= \mathcal{L}_{fm} + \lambda_{rec}\mathcal{L}_{rec} +\lambda_{id}\mathcal{L}_{id},
    \label{eq:rec}
\end{align}
where $\lambda_{rec}$, $\lambda_{id}$, and $\lambda_{gp}$ are the hyper-parameters for respective loss terms. More implementation details can be found in \cref{sec:implementation-details}.

\section{Experiments}
In this section, we introduce the experimental setup and showcase the efficacy of our framework in a series of experiments. Our method is evaluated both qualitatively and quantitatively and compared against two strong baselines (\cref{section:baselines}). The main focus of our quantitative experiments are: a) aging accuracy (\cref{section:aging-accuracy}) and b) enhancement of diversity (\cref{section:diversity}).

\subsection{Implementation details}
\label{sec:implementation-details}
For all the experiments, both the encoder and decoder of $\mathcal{G}$, as well as $\mathcal{D}$ have 6 layers. The exact architectures are analyzed in the supplementary material. We utilize skip connections between all layers of the encoder and the decoder. 
To improve the stability of the training we include the $R_1$ gradient penalty objective in addition to the adversarial loss for the discriminator \cite{mescheder_which_2018}, which is defined as:
\begin{equation}
   \mathcal{L}_{gp} = \lambda_{gp} \mathbb{E}_\mathbf{x}\left[ \parallel \nabla \mathcal{D}(\mathbf{x}) \parallel^2 \right].
\end{equation}
Furthermore, instead of using the original image as the target in the cycle loss (\cref{eq:rec}), we find it beneficial to use the translated target image $\mathbf{\tilde{x}}'_A = \mathcal{G}(\mathbf{x}'_B, \mathbf{x}_A)$. That is, instead of allowing the input image to drive the reconstruction, we utilize a different image with the age-specific style of the input. By doing this, we further enforce the transfer of the aging features through reconstruction. The model is trained end-to-end with hyperparameters $\lambda_{rec}=0.01$, $\lambda_{gp}=10.0$, and $\lambda_{id}=10^{-4}$. The networks' weights are optimised with Adam \cite{adam}, with a learning rate of $10^{-4}$, and beta values $\beta_1=0.5,\beta_2=0.99$. All images are aligned and resized to $128\times128$.
\begin{table*}[h!]
\centering    
\begin{tabular}{cccc||ccc}

\hline
&\multicolumn{3}{c}{MOPRH} &\multicolumn{3}{c}{CACD}\\
\hline
&31-40&41-50&50+&31-40&41-50&50+\\
\hline

GT &$35.9\pm2.65$&$44.77\pm2.72$&$54.92\pm3.72$ & $35.41\pm2.88$&$45.45\pm2.88$&$55.01\pm3.02$  \\ 
\hline
CAAE &$37.08\pm4.53$&$39.25\pm4.54$&$41.96\pm4.68$& $38.47\pm5.46$&$41.38\pm5.17$&$43.30\pm5.49$  \\
IPCGAN &$41.86\pm6.71$&$47.94\pm8.47$&$50.89\pm6.35$ & $35.37\pm7.09$&$42.25\pm8.13$&$40.79\pm7.43$  \\
Ours &$39.17\pm6.46$&$45.98\pm6.00$&$56.62\pm5.35$ &$33.04\pm7.40$&$46.78\pm7.00$&$56.05\pm6.06$  \\
\hline
&\multicolumn{6}{c}{Mean absolute age difference between synthetic images and GT (years)}\\
\hline
CAAE &\textbf{1.18}&5.52&12.96&3.06&4.07&11.71  \\
IPCGAN &5.96&3.17&4.03&\textbf{0.04}&3.20&14.22  \\
Ours &3.27&\textbf{1.21}&\textbf{1.69}&2.37&\textbf{1.33}&\textbf{1.04}  \\
\hline

\end{tabular}

\caption{Age accuracy of the proposed method and baseline models on the test sets of MORPH and CACD. We translate images from the $-30$ group to all other age groups. `GT' is the mean age and standard deviation of the ground-truth test images. The estimated ages for all models are obtained using DEX \cite{dexRothe-ICCVW-2015} on the generated images.}
\label{tab:age_comparison}
\end{table*}

\begin{table*}[hbt!]
\centering    
\begin{tabular}{ccccc||cccc||cccc}

\hline
&\multicolumn{4}{c}{MOPRH} &\multicolumn{4}{c}{CACD}&\multicolumn{4}{c}{FG-NET}\\
\hline
&ShH&ShE&SiD&SiE&ShH&ShE&SiD&SiE&ShH&ShE&SiD&SiE\\
\hline

GT &1.17&0.85&2.89&0.72 &1.34&0.97&3.69&0.92& 1.05&0.75&2.32&0.58\\
\hline
CAAE &1.1&0.79&2.62&0.65 &1.20&0.87&2.95&0.74 &1.26&0.91&3.29&0.82\\
IPCGAN &1.31&0.95&3.49&0.87 &1.32&0.95&3.56&0.89 &1.33&0.96&3.59&0.90  \\
Ours &\textbf{1.36}&\textbf{0.98}&\textbf{3.81}&\textbf{0.95} &\textbf{1.35}&\textbf{0.97}&\textbf{3.75}&\textbf{0.94} &\textbf{1.36}&\textbf{0.97}&\textbf{3.72}&\textbf{0.93} \\
\hline

\end{tabular}

\caption{Diversity metrics of the augmented test sets. The proposed method outperforms the baselines on all datasets. `GT' denotes the diversity indeces of the original test sets.}
\label{tab:diversity}
\end{table*}

\subsection{Datasets}
\label{section:datasets}

We benchmark our model using the \textbf{MORPH} \cite{morph} and \textbf{CACD} \cite{cacd} datasets. The second album of MORPH contains over 55,134 images of 13,618 people. Most images are near-frontal and the capture conditions (e.g., background and illumination) are almost uniform. The age of the faces in MORPH range from 16 to 77 years old. On the other hand, the CACD dataset  consists of over 160,000 images from 2,000 celebrities. The images are collected from Google Images and are hence captured in-the-wild. The age of the subjects ranges from 14 to 62 years old. For both datasets, we use 20\% of the images for testing and keep 80\% for training the models. The generalization of the models is tested on \textbf{FG-NET} \cite{FGNET}, which has 1,002 face images of 82 subjects. Following the standard approach (\cite{psdganYang_2018, wang2016recurrent, hfa}) we utilize 4 age groups: under 30, 30-40, 40-50, and over 50 years old.

\subsection{Baselines}
\label{section:baselines}

We compare our method with two recent age progression methods, namely \textbf{CAAE} \cite{caae} and \textbf{IPCGAN} \cite{ipcgan}. CAAE performs age progression and regression by traversing a low-dimensional manifold. On the other hand, IPCGAN utilizes two external pre-trained networks that capture the identity and age of the synthesized face. The baseline models are compared to the proposed framework in a series of experiments. Both qualitative and quantitative results are presented in the sections that follow. Both models were trained using the authors' provided source code\footnote{\textbf{CAAE}:
\href{https://github.com/ZZUTK/Face-Aging-CAAE and}{https://github.com/ZZUTK/Face-Aging-CAAE} and\\ \textbf{IPCGAN}: \href{https://github.com/dawei6875797/Face-Aging-with-Identity-Preserved-Conditional-Generative-Adversarial-Networks}{https://github.com/dawei6875797/Face-Aging-with-Identity-Preserved-Conditional-Generative-Adversarial-Networks}}.

\subsection{Qualitative Results}
\label{section:qualitative-results}

We present the results for age transfer on the test sets of MOPRH and CACD in \cref{fig:qual-compare} (additional results are included in the supplementary material). Despite the variation in capturing conditions, gender and facial expression of the datasets, our model is able to produce realistic aged and rejuvenated renderings of the input. In particular, each input face is translated to the 3 remaining age classes (except for the ground-truth) using the proposed framework, as well as the baseline methods. We notice that CAAE generates relatively blurry and over-regularized faces, that do not always maintain the identity of the input. On the other hand, while IPCGAN is able to produce sharp images of the target age group, it nevertheless fails to synthesize convincingly old (over 50) and young (under 30) faces, which is crucial for mitigating age bias. The proposed method is able to generate both young and old-looking faces by transferring aging patterns such as wrinkles and hair color. Additionally, the proposed method is able to produce more diverse aging patterns for faces aged 31-40 and 41-50 years old, whereas the baseline models synthesize only subtle changes between these two adjacent groups. Lastly, in order to test the generalisation of the methods, we test the models on the entire unseen FG-NET dataset, using the models trained on CACD. The results in \cref{fig:qual-compare}c are consistent with the above, with our model generating more photo-realistic aged faces.

\begin{figure*}
    \centering
    \includegraphics[width=0.9\textwidth]{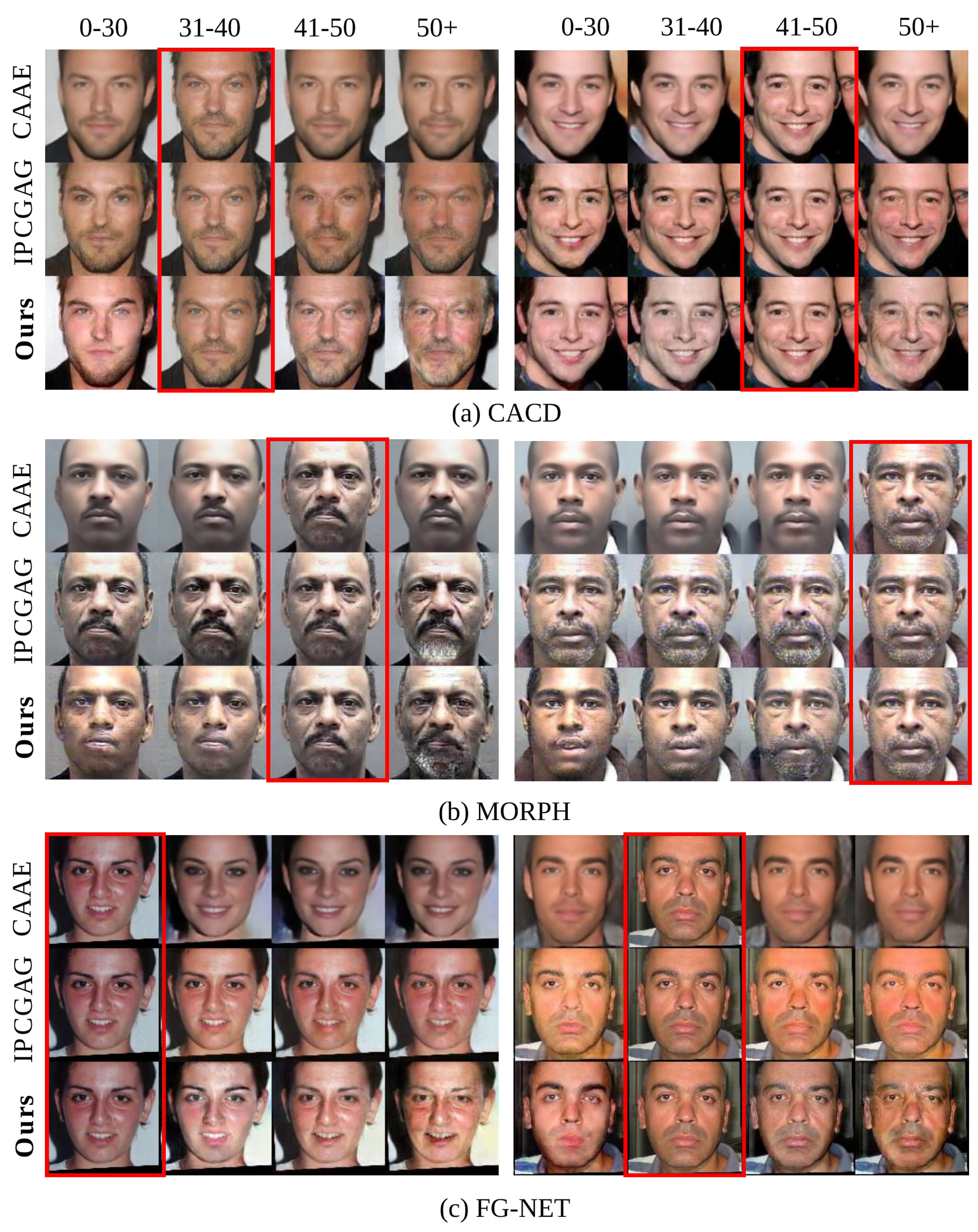}
    \caption{Samples generated by the proposed and  baseline methods. Each input image from the test set is translated to the remaining age groups. The images in the red rectangle are the input faces and are positioned in their corresponding age group's column.}
    \label{fig:qual-compare}
\end{figure*}

\subsection{Aging accuracy}
\label{section:aging-accuracy}

The purpose of age transfer is to translate an input image so that it presents the age features of a target age group. In this section we evaluate the accuracy of these age features by using a pre-trained age estimation network \cite{dexRothe-ICCVW-2015}. In particular, we perform age progression on faces under 30 years old and use the age estimation model to obtain the mean and standard deviation of the predicted ages. The estimated age of the synthetic images should follow the distribution of the real images, hence we evaluate the models based on the deviation between the mean age of the synthetic and the real images for each age group.  Aging accuracy results for all compared methods are presented in \cref{tab:age_comparison}. 
We observe that CAAE consistently produces similar age patterns, while IPCGAN is not able to generate sufficiently old-looking faces over 50 years old.

\subsection{Diversity enhancement}
\label{section:diversity}

In this section, we quantify the ability of our model to enhance the diversity of a dataset. In particular, we measure the Shannon H (ShH) and E (ShE) and the Simpson D (SiD) and E (SiE) indices, as proposed in \cite{dif-merler2019diversity}. Simpson D and Shannon H measure the diversity  of  the  dataset, while  Simpson  E  and  Shannon  E  quantify  the  evenness  of  the  distribution. The indices are calculated as follows:

\begin{alignat}{3}
    &Shannon: \; \; &&H = - \sum_1^Sp_i\; \ln(p_i), \quad &&E = \frac{H}{\ln(S)} \nonumber\\
    &Simpson: &&D = \frac{1}{\sum_1^Sp_i^2} , &&E = \frac{D}{S}, \nonumber
\end{alignat}
where $S$ denotes the number of classes and $p_i$ is the probability of each class. In general, larger values of Simpson D and Shannon H indicate a more diverse dataset, while Simpson  E  and  Shannon  E closer to 1 indicate a more even distribution. We focus only on the age distribution of a dataset and measure the diversity indices for MORPH, CACD and FG-NET. The results on \cref{tab:diversity} indicate the imbalanced distribution of MORPH (only 7\% of the test set are over 50 years old) and FG-NET (87\% of the  faces are under 30 years old).

In order to benchmark the diversity enhancing capabilities of the proposed framework and the baseline models, the datasets are augmented using all 3 methods. Each face in the test sets is translated to the remaining 3 age classes, resulting in an augmented dataset that is 4 times the size of the original dataset. We subsequently measure the diversity indices for the augmented datasets and report the results in \cref{tab:diversity}. The results indicate that only the proposed method is able to generate a distribution of ages that is almost even. On the contrary, the inability of CAAE to generate significant facial transformations deteriorates the diversity of the datasets significantly.

\subsection{Diversity in aging patterns}
\label{section:intraclass}
\begin{figure}[h]
\centering
    \includegraphics[width=1.0\linewidth]{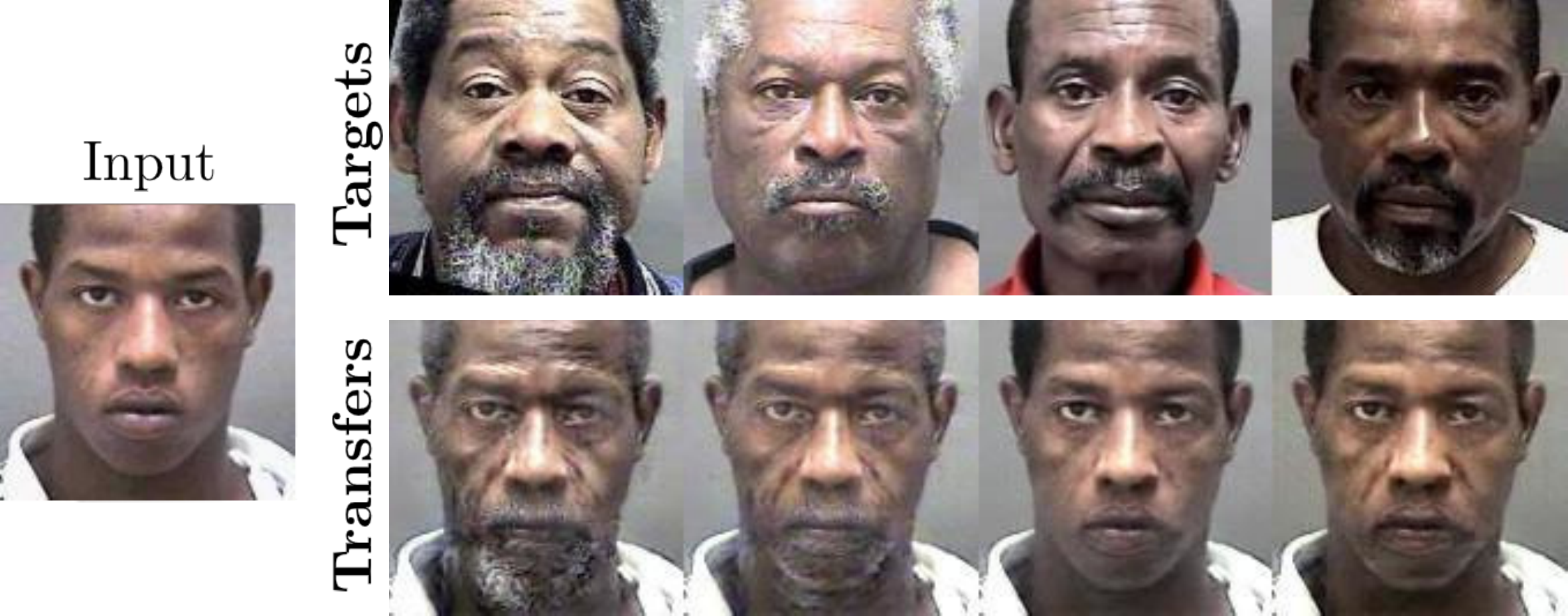}
    \caption{Age progression of an input face using different targets (top row). By conditioning the age transfer on different targets, we are able to synthesize different age-specific facial features.}
    \label{fig:diversity}
\end{figure}

In this work, we introduce an approach to age progression that is different to the standard paradigm. In particular, the proposed method transfers the age-discriminative style of a target face onto the input face at multiple scales. This approach allows for the generation of diverse aging patterns, based on the choice of target image. This is demonstrated in \cref{fig:diversity}, where a young (under 30 years old) input face is aged using different target faces over 50 years old. It is evident that different aging patterns (e.g., white hair, beard, and wrinkles) are transferred according to the target.

\begin{figure*}
    \centering
    \includegraphics[width=1.0\textwidth]{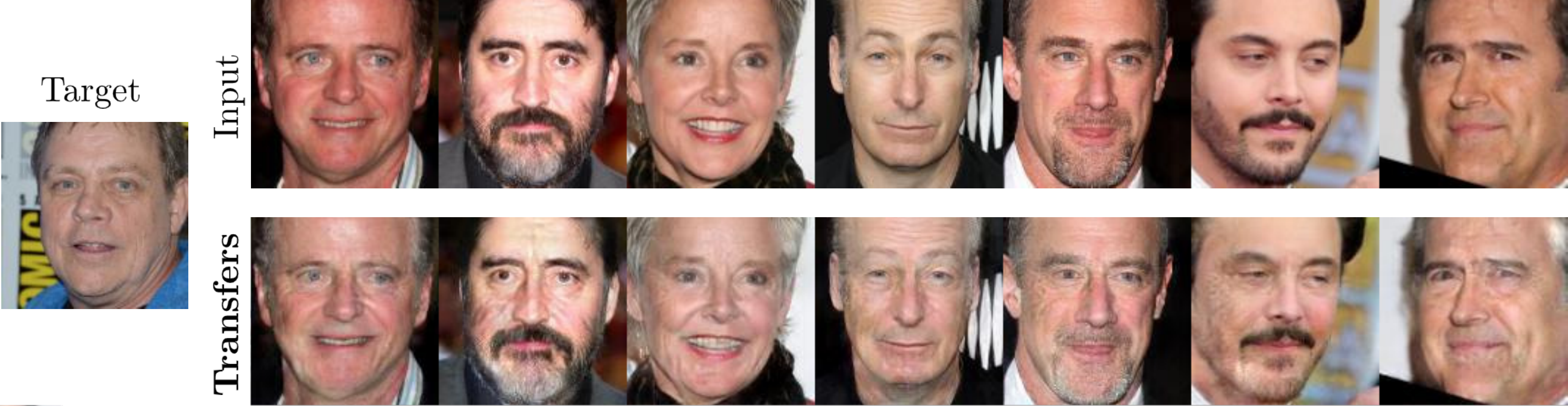}
    \caption{Age progressions of faces over 50 years old. Each input image (top row) is over 50 years old and is translated to the same age group using an older target image. The model is able to transfer more pronounced aging patterns and synthesize older looking faces of the same age group.} 
    \label{fig:twos}
\end{figure*}

The ability to transfer diverse aging patterns is vital, especially for the case of celebrity datasets (e.g., CACD). That is, celebrity faces do not display the same aging patterns as non-celebrity ones and tend to look younger. This affects the performance of age progression as shown in \cref{fig:qual-compare}a, where none of the baselines are able to generate sufficiently old-looking faces over 50 years old. In \cref{fig:twos} we also demonstrate how our model is able to mitigate the apparent age bias of celebrity faces. In particular, we transfer more crude aging features to faces over 50 years old. The resulting faces look significantly older and hence, can be used to enhance the diversity of the dataset.

\section{Conclusion}

In this paper, we introduce a novel face aging method to enhance the diversity with respect to age in facial datasets. Inspired by the style transfer literature, the proposed method is able to transfer the aging patterns of a target image. We demonstrate the ability of our model to generate realistic age progressions in a series of quantitative and qualitative experiments. 
Furthermore, we propose to benchmark the efficacy of the evaluated models in enhancing dataset diversity using the metrics proposed in \cite{dif-merler2019diversity}. The proposed method outperforms the baselines and is able to generate even age distributions and mitigate the dataset bias. As a future direction, we plan to generalize the proposed framework to multiple demographic attributes, e.g., gender and race.

{\small
\bibliographystyle{ieee_fullname}
\bibliography{}

\begin{thebibliography}{10}\itemsep=-1pt

\bibitem{alvi2018turning}
Mohsan Alvi, Andrew Zisserman, and Christoffer Nell{\aa}ker.
\newblock Turning a blind eye: Explicit removal of biases and variation from
  deep neural network embeddings.
\newblock In {\em Proceedings of the European Conference on Computer Vision
  (ECCV)}, pages 0--0, 2018.

\bibitem{burt1995perception}
D~Michael Burt and David~I Perrett.
\newblock Perception of age in adult caucasian male faces: Computer graphic
  manipulation of shape and colour information.
\newblock {\em Proceedings of the Royal Society of London. Series B: Biological
  Sciences}, 259(1355):137--143, 1995.

\bibitem{cacd}
Bor-Chun Chen, Chu-Song Chen, and Winston~H. Hsu.
\newblock Cross-age reference coding for age-invariant face recognition and
  retrieval.
\newblock In {\em Proceedings of the European Conference on Computer Vision
  ({ECCV})}, 2014.

\bibitem{captionChen_2019}
Chen Chen, Shuai Mu, Wanpeng Xiao, Zexiong Ye, Liesi Wu, and Qi Ju.
\newblock Improving image captioning with conditional generative adversarial
  nets.
\newblock {\em Proceedings of the AAAI Conference on Artificial Intelligence},
  33:8142–8150, Jul 2019.

\bibitem{wganresolution-chen2017face}
Zhimin Chen and Yuguang Tong.
\newblock Face super-resolution through wasserstein gans, 2017.

\bibitem{starganChoi_2018}
Yunjey Choi, Minje Choi, Munyoung Kim, Jung-Woo Ha, Sunghun Kim, and Jaegul
  Choo.
\newblock Stargan: Unified generative adversarial networks for multi-domain
  image-to-image translation.
\newblock {\em 2018 IEEE/CVF Conference on Computer Vision and Pattern
  Recognition}, Jun 2018.

\bibitem{naturalcaptionDai_2017}
Bo Dai, Sanja Fidler, Raquel Urtasun, and Dahua Lin.
\newblock Towards diverse and natural image descriptions via a conditional gan.
\newblock {\em 2017 IEEE International Conference on Computer Vision (ICCV)},
  Oct 2017.

\bibitem{fu2010age}
Yun Fu, Guodong Guo, and Thomas~S Huang.
\newblock Age synthesis and estimation via faces: A survey.
\newblock {\em IEEE transactions on pattern analysis and machine intelligence},
  32(11):1955--1976, 2010.

\bibitem{gatys}
Leon~A. Gatys, Alexander~S. Ecker, and Matthias Bethge.
\newblock A {Neural} {Algorithm} of {Artistic} {Style}.
\newblock {\em arXiv:1508.06576 [cs, q-bio]}, Sept. 2015.
\newblock arXiv: 1508.06576.

\bibitem{georgopoulos2018modeling}
Markos Georgopoulos, Yannis Panagakis, and Maja Pantic.
\newblock Modeling of facial aging and kinship: A survey.
\newblock {\em Image and Vision Computing}, 80:58--79, 2018.

\bibitem{Gu2018ArbitraryST}
Shuyang Gu, Congliang Chen, Jing Liao, and Lu Yuan.
\newblock Arbitrary style transfer with deep feature reshuffle.
\newblock {\em 2018 IEEE/CVF Conference on Computer Vision and Pattern
  Recognition}, pages 8222--8231, 2018.

\bibitem{adain}
Xun Huang and Serge Belongie.
\newblock Arbitrary {Style} {Transfer} in {Real}-time with {Adaptive}
  {Instance} {Normalization}.
\newblock {\em arXiv:1703.06868 [cs]}, July 2017.
\newblock arXiv: 1703.06868.

\bibitem{munit}
Xun Huang, Ming-Yu Liu, Serge Belongie, and Jan Kautz.
\newblock Multimodal {Unsupervised} {Image}-to-{Image} {Translation}.
\newblock {\em arXiv:1804.04732 [cs, stat]}, Aug. 2018.
\newblock arXiv: 1804.04732.

\bibitem{pix2pixIsola_2017}
Phillip Isola, Jun-Yan Zhu, Tinghui Zhou, and Alexei~A. Efros.
\newblock Image-to-image translation with conditional adversarial networks.
\newblock {\em 2017 IEEE Conference on Computer Vision and Pattern Recognition
  (CVPR)}, Jul 2017.

\bibitem{Johnson_2016}
Justin Johnson, Alexandre Alahi, and Li Fei-Fei.
\newblock Perceptual losses for real-time style transfer and super-resolution.
\newblock {\em Lecture Notes in Computer Science}, page 694–711, 2016.

\bibitem{stylegan}
Tero Karras, Samuli Laine, and Timo Aila.
\newblock A {Style}-{Based} {Generator} {Architecture} for {Generative}
  {Adversarial} {Networks}.
\newblock {\em arXiv:1812.04948 [cs, stat]}, Mar. 2019.
\newblock arXiv: 1812.04948.

\bibitem{kemelmacher2014illumination}
Ira Kemelmacher-Shlizerman, Supasorn Suwajanakorn, and Steven~M Seitz.
\newblock Illumination-aware age progression.
\newblock In {\em Proceedings of the IEEE conference on computer vision and
  pattern recognition}, pages 3334--3341, 2014.

\bibitem{kim2019learning}
Byungju Kim, Hyunwoo Kim, Kyungsu Kim, Sungjin Kim, and Junmo Kim.
\newblock Learning not to learn: Training deep neural networks with biased
  data.
\newblock In {\em Proceedings of the IEEE Conference on Computer Vision and
  Pattern Recognition}, pages 9012--9020, 2019.

\bibitem{ugat}
Junho Kim, Minjae Kim, Hyeonwoo Kang, and Kwanghee Lee.
\newblock U-{GAT}-{IT}: {Unsupervised} {Generative} {Attentional} {Networks}
  with {Adaptive} {Layer}-{Instance} {Normalization} for {Image}-to-{Image}
  {Translation}.
\newblock {\em arXiv:1907.10830 [cs, eess]}, Jan. 2020.
\newblock arXiv: 1907.10830.

\bibitem{adam}
Diederik Kingma and Jimmy Ba.
\newblock Adam: A method for stochastic optimization.
\newblock {\em International Conference on Learning Representations}, 12 2014.

\bibitem{FGNET}
A. Lanitis.
\newblock F{G-NET} {A}ging {D}atabase.
\newblock 2002.

\bibitem{superresolutionLedig_2017}
Christian Ledig, Lucas Theis, Ferenc Huszar, Jose Caballero, Andrew Cunningham,
  Alejandro Acosta, Andrew Aitken, Alykhan Tejani, Johannes Totz, Zehan Wang,
  and et al.
\newblock Photo-realistic single image super-resolution using a generative
  adversarial network.
\newblock {\em 2017 IEEE Conference on Computer Vision and Pattern Recognition
  (CVPR)}, Jul 2017.

\bibitem{Li_2016}
Chuan Li and Michael Wand.
\newblock Precomputed real-time texture synthesis with markovian generative
  adversarial networks.
\newblock {\em Lecture Notes in Computer Science}, page 702–716, 2016.

\bibitem{universal-wct}
Yijun Li, Chen Fang, Jimei Yang, Zhaowen Wang, Xin Lu, and Ming-Hsuan Yang.
\newblock Universal style transfer via feature transforms, 2017.

\bibitem{demyst}
Yanghao Li, Naiyan Wang, Jiaying Liu, and Xiaodi Hou.
\newblock Demystifying neural style transfer.
\newblock {\em Proceedings of the Twenty-Sixth International Joint Conference
  on Artificial Intelligence}, Aug 2017.

\bibitem{batchnorm-domain}
Yanghao Li, Naiyan Wang, Jianping Shi, Jiaying Liu, and Xiaodi Hou.
\newblock Revisiting batch normalization for practical domain adaptation, 2016.

\bibitem{fewshot}
Ming-Yu Liu, Xun Huang, Arun Mallya, Tero Karras, Timo Aila, Jaakko Lehtinen,
  and Jan Kautz.
\newblock Few-{Shot} {Unsupervised} {Image}-to-{Image} {Translation}.
\newblock May 2019.

\bibitem{closed-form-optimal-transport}
Ming Lu, Hao Zhao, Anbang Yao, Yurong Chen, Feng Xu, and Li Zhang.
\newblock A closed-form solution to universal style transfer.
\newblock {\em 2019 IEEE/CVF International Conference on Computer Vision
  (ICCV)}, Oct 2019.

\bibitem{dif-merler2019diversity}
Michele Merler, Nalini Ratha, Rogerio~S. Feris, and John~R. Smith.
\newblock Diversity in faces, 2019.

\bibitem{mescheder_which_2018}
Lars Mescheder, Andreas Geiger, and Sebastian Nowozin.
\newblock Which {Training} {Methods} for {GANs} do actually {Converge}?
\newblock {\em arXiv:1801.04406 [cs]}, July 2018.
\newblock arXiv: 1801.04406.

\bibitem{agedb}
Stylianos Moschoglou, Athanasios Papaioannou, Christos Sagonas, Jiankang Deng,
  Irene Kotsia, and Stefanos Zafeiriou.
\newblock Agedb: the first manually collected, in-the-wild age database.
\newblock In {\em Proceedings of the IEEE Conference on Computer Vision and
  Pattern Recognition Workshop}, volume~2, page~5, 2017.

\bibitem{mroueh2019wasserstein}
Youssef Mroueh.
\newblock Wasserstein style transfer, 2019.

\bibitem{afad}
Z. {Niu}, M. {Zhou}, L. {Wang}, X. {Gao}, and G. {Hua}.
\newblock Ordinal regression with multiple output cnn for age estimation.
\newblock In {\em 2016 IEEE Conference on Computer Vision and Pattern
  Recognition (CVPR)}, pages 4920--4928, 2016.

\bibitem{quadrianto2018discovering}
Novi Quadrianto, Viktoriia Sharmanska, and Oliver Thomas.
\newblock Discovering fair representations in the data domain, 2018.

\bibitem{dcganradford2015unsupervised}
Alec Radford, Luke Metz, and Soumith Chintala.
\newblock Unsupervised representation learning with deep convolutional
  generative adversarial networks, 2015.

\bibitem{ramanathan2006modeling}
Narayanan Ramanathan and Rama Chellappa.
\newblock Modeling age progression in young faces.
\newblock In {\em 2006 IEEE Computer Society Conference on Computer Vision and
  Pattern Recognition (CVPR'06)}, volume~1, pages 387--394. IEEE, 2006.

\bibitem{ramanathan2009age}
Narayanan Ramanathan, Rama Chellappa, Soma Biswas, et~al.
\newblock Age progression in human faces: A survey.

\bibitem{morph}
K. {Ricanek} and T. {Tesafaye}.
\newblock Morph: a longitudinal image database of normal adult age-progression.
\newblock In {\em 7th International Conference on Automatic Face and Gesture
  Recognition (FGR06)}, pages 341--345, April 2006.

\bibitem{unetRonneberger_2015}
Olaf Ronneberger, Philipp Fischer, and Thomas Brox.
\newblock U-net: Convolutional networks for biomedical image segmentation.
\newblock {\em Medical Image Computing and Computer-Assisted Intervention –
  MICCAI 2015}, page 234–241, 2015.

\bibitem{dexRothe-ICCVW-2015}
Rasmus Rothe, Radu Timofte, and Luc~Van Gool.
\newblock Dex: Deep expectation of apparent age from a single image.
\newblock In {\em IEEE International Conference on Computer Vision Workshops
  (ICCVW)}, December 2015.

\bibitem{improved-gans}
Tim Salimans, Ian~J. Goodfellow, Wojciech Zaremba, Vicki Cheung, Alec Radford,
  and Xi Chen.
\newblock Improved techniques for training gans.
\newblock {\em CoRR}, abs/1606.03498, 2016.

\bibitem{Sanakoyeu_2018}
Artsiom Sanakoyeu, Dmytro Kotovenko, Sabine Lang, and Björn Ommer.
\newblock A style-aware content loss for real-time hd style transfer.
\newblock {\em Lecture Notes in Computer Science}, page 715–731, 2018.

\bibitem{sattigeri2018fairness}
Prasanna Sattigeri, Samuel~C. Hoffman, Vijil Chenthamarakshan, and Kush~R.
  Varshney.
\newblock Fairness gan, 2018.

\bibitem{suo2009compositional}
Jinli Suo, Song-Chun Zhu, Shiguang Shan, and Xilin Chen.
\newblock A compositional and dynamic model for face aging.
\newblock {\em IEEE Transactions on Pattern Analysis and Machine Intelligence},
  32(3):385--401, 2009.

\bibitem{todd1980perception}
James~T Todd, Leonard~S Mark, Robert~E Shaw, and John~B Pittenger.
\newblock The perception of human growth.
\newblock {\em Scientific american}, 242(2):132--145, 1980.

\bibitem{torralba2011unbiased}
Antonio Torralba and Alexei~A Efros.
\newblock Unbiased look at dataset bias.
\newblock In {\em CVPR 2011}, pages 1521--1528. IEEE, 2011.

\bibitem{ulyanov2016texture}
Dmitry Ulyanov, Vadim Lebedev, Andrea Vedaldi, and Victor Lempitsky.
\newblock Texture networks: Feed-forward synthesis of textures and stylized
  images, 2016.

\bibitem{wang2016recurrent}
Wei Wang, Zhen Cui, Yan Yan, Jiashi Feng, Shuicheng Yan, Xiangbo Shu, and Nicu
  Sebe.
\newblock Recurrent face aging.
\newblock In {\em Proceedings of the IEEE Conference on Computer Vision and
  Pattern Recognition}, pages 2378--2386, 2016.

\bibitem{ipcgan}
Z. Wang, W.~Luo X.~Tang, and S. Gao.
\newblock Face aging with identity-preserved conditional generative adversarial
  networks.
\newblock In {\em 2018 IEEE Conference on Computer Vision and Pattern
  Recognition (CVPR)}, 2018.

\bibitem{fairganXu_2018}
Depeng Xu, Shuhan Yuan, Lu Zhang, and Xintao Wu.
\newblock Fairgan: Fairness-aware generative adversarial networks.
\newblock {\em 2018 IEEE International Conference on Big Data (Big Data)}, Dec
  2018.

\bibitem{psdganYang_2018}
Hongyu Yang, Di Huang, Yunhong Wang, and Anil~K. Jain.
\newblock Learning face age progression: A pyramid architecture of gans.
\newblock {\em 2018 IEEE/CVF Conference on Computer Vision and Pattern
  Recognition}, Jun 2018.

\bibitem{hfa}
Hongyu Yang, Di Huang, Yunhong Wang, Heng Wang, and Yuanyan Tang.
\newblock Face aging effect simulation using hidden factor analysis joint
  sparse representation.
\newblock {\em IEEE Transactions on Image Processing}, 25(6):2493--2507, 2016.

\bibitem{caae}
Zhifei Zhang, Yang Song, and Hairong Qi.
\newblock Age progression/regression by conditional adversarial autoencoder.
\newblock In {\em IEEE Conference on Computer Vision and Pattern Recognition
  (CVPR)}, 2017.

\bibitem{cycleganZhu_2017}
Jun-Yan Zhu, Taesung Park, Phillip Isola, and Alexei~A. Efros.
\newblock Unpaired image-to-image translation using cycle-consistent
  adversarial networks.
\newblock {\em 2017 IEEE International Conference on Computer Vision (ICCV)},
  Oct 2017.

\end{thebibliography}
}

\end{document}